\documentclass[9pt,conference]{IEEEtran}
\IEEEoverridecommandlockouts
\usepackage{cite}
\usepackage{amsmath,amssymb,amsfonts}
\usepackage{algorithmic}
\usepackage{graphicx}
\usepackage{textcomp}
\usepackage{xcolor}
\usepackage{subcaption}

\usepackage{cite}
\usepackage{amsmath,amssymb,amsfonts}
\usepackage{algorithmic}
\usepackage{graphicx}
\usepackage{textcomp}

\usepackage[hidelinks]{hyperref}
\usepackage{xcolor}

\usepackage{amsmath,amssymb,amsfonts}
\usepackage{amssymb}
\usepackage{algorithmic}
\usepackage{graphicx}
\usepackage{textcomp}
\usepackage{xcolor}
\usepackage{subcaption}

\usepackage{color,xcolor}
\usepackage{epsfig}
\usepackage{graphicx}
\usepackage{verbatim}

\usepackage{adjustbox}
\usepackage{array}
\usepackage{booktabs}
\usepackage{multirow}

\usepackage{amsmath,amsfonts,amsthm,amssymb}
\usepackage{bm}
\usepackage{nicefrac}
\usepackage{microtype}

\usepackage{changepage}
\usepackage{extramarks}
\usepackage{fancyhdr}
\usepackage{lastpage}
\usepackage{setspace}
\usepackage{soul}
\usepackage{xspace}

\usepackage{marvosym}
\usepackage{comment}

\newcommand\norm[1]{\left\lVert#1\right\rVert}

\def\BibTeX{{\rm B\kern-.05em{\sc i\kern-.025em b}\kern-.08em
    T\kern-.1667em\lower.7ex\hbox{E}\kern-.125emX}}
\begin{document}

\title{Federated Contrastive Learning and Masked Autoencoder for Dermatological Disease Diagnosis\thanks{The original conference version of this work was published on ICCAD 2021 proceedings \cite{wu2021federated}. Compared with the original version, major extensions have been made in this manuscript.}}

\author{Yawen Wu$^{1}$, Dewen Zeng$^{2}$, Zhepeng Wang$^{3}$, Yi Sheng$^{3}$, Lei Yang$^{3}$, Alaina J. James$^{4}$, Yiyu Shi$^{2}$, and Jingtong Hu$^{1}$ \\
$^{1}$Department of Electrical and Computer Engineering, University of Pittsburgh, USA \\
$^{2}$Department of Computer Science and Engineering, University of Notre Dame, USA \\
$^{3}$Department of Electrical and Computer Engineering, George Mason University, USA \\
$^{4}$University of Pittsburgh Medical Center, USA \\
Email: yawen.wu@pitt.edu, dzeng2@nd.edu, zwang48@gmu.edu, ysheng2@gmu.edu, \\ lyang29@gmu.edu, jamesaj@upmc.edu, yshi4@nd.edu, jthu@pitt.edu}

\maketitle

\begin{abstract}
In dermatological disease diagnosis, the private data collected by mobile dermatology assistants exist on distributed mobile devices of patients.
Federated learning (FL) can use decentralized data to train models while keeping data local.
Existing FL methods assume all the data have labels.
However, medical data often comes without full labels due to high labeling costs.
Self-supervised learning (SSL) methods, contrastive learning (CL) and masked autoencoders (MAE), can leverage the unlabeled data to pre-train models, followed by fine-tuning with limited labels.
However, combining SSL and FL has unique challenges. 
For example, CL requires diverse data but each device only has limited data.
For MAE, 
while Vision Transformer (ViT) based MAE has higher accuracy over CNNs in centralized learning, MAE's performance in FL with unlabeled data has not been investigated.
Besides, the ViT synchronization between the server and clients is different from traditional CNNs. Therefore, special synchronization methods need to be designed.
In this work, we propose two federated self-supervised learning frameworks for dermatological disease diagnosis with limited labels. The first one features lower computation costs, suitable for mobile devices. The second one features high accuracy and fits high-performance servers.
Based on CL, we proposed federated contrastive learning with feature sharing (FedCLF). Features are shared for diverse contrastive information without sharing raw data for privacy. 
Based on MAE, we proposed FedMAE. Knowledge split separates the global and local knowledge learned from each client. Only global knowledge is aggregated for higher generalization performance.
Experiments on dermatological disease datasets show superior accuracy of the proposed frameworks over state-of-the-arts.

\end{abstract}

\section{Introduction}

Skin diseases are a major global health threat to a tremendous amount of people in the world \cite{verma2019classification}.
These diseases not only injure the physical health including the risk of
skin cancer but also can result in psychological problems such as lack of self-confidence and psychological depression due to damaged appearance \cite{ahmad2020discriminative, wu2019studies}.
Deep learning models have shown great promise in skin disease diagnosis \cite{wu2019studies, velasco2019smartphone,gu2019progressive,wu2022fairprune} and have been widely deployed on mobile devices as mobile dermatology assistants \cite{googleai, firstderm, DermExpert}.
These models are trained on a large amount of data with full labels to achieve a high accuracy \cite{wu2020enabling}. 
When the large-scale datasets for training are not available, the performance of deep learning models will greatly degrade \cite{kairouz2019advances}.
However, the images of skin disease are usually distributed on mobile devices of patients, which are impractical and even illegal to combine in a single location \cite{kairouz2019advances} to form large-scale datasets since data sharing is constrained by the Health Insurance Portability and Accountability Act (HIPAA) \cite{kairouz2019advances}.
For example, skin disease images can be taken by the cameras of mobile devices and stored for a preliminary self-diagnosis \cite{velasco2019smartphone,sun2016benchmark}.
But patients are usually reluctant to share highly private and sensitive images with the data center. 
Without large-scale datasets in a single location, it is not feasible to perform centralized training for learning an accurate model.

Federated learning (FL) is a distributed learning framework where many mobile devices collaboratively learn a global prediction model without sharing private data \cite{yang2019federated}. 
By leveraging FL, distributed data on mobile devices 
can be used to train an accurate shared model to diagnose skin diseases while keeping data local.
Existing FL works assume the local data on devices are fully labeled and use supervised learning for local model updates. However, the assumption of fully labeled data is impractical. For instance, the patients may not want to spend time labeling their skin images captured by cameras on mobile phones. Even voluntary patients may not be able to accurately label all their own images due to the lack of expertise. 
Therefore, most of the distributed data on devices will be unlabeled, which makes supervised FL unrealistic.

Recently developed self-supervised learning (SSL) approaches have advanced the visual representation learning from unlabeled data and effectively improves the model accuracy on the downstream task such as disease diagnosis.
The state-of-the-art SSL methods can be categorized as contrastive learning and generative learning, both of which have the potential to be combined with FL to leverage the decentralized unlabeled data for more accurate disease diagnosis.
These two categories of SSL methods have different model performance and computation costs. 
Based on them, we propose two frameworks to enable federated self-supervised learning. 
The first framework is based on contrastive learning with CNN backbones and it features lower computation costs compared with the second one. This framework is suitable for computation-constrained mobile devices.
The second framework is based on generative learning and Vision Transformers backbones, which features higher accuracy with higher computation cost. This framework is suitable for distributed medical institutions with high-performance computation devices such as server GPUs.

The \textbf{first} SSL category, contrastive learning (CL), is a recently developed technique to learn effective visual representations of data without using labels \cite{he2020momentum}. 
CL learns by comparing different views of images. 
By combining CL with FL as federated contrastive learning, the conventional supervised learning on local devices can be replaced by CL pre-training without using labels.
After that, the pre-trained model can be used as the initialization to fine-tune for the diagnosis task with limited labels.
In this way, an accurate dermatological disease diagnosis model can be learned by using distributed data with limited labels.

However, simply combining CL into FL cannot achieve optimal performance. 
This is because existing CL approaches \cite{he2020momentum, chen2020simple, caron2020unsupervised} are originally developed for centralized training on large-scale datasets, assuming sufficiently diverse data is available for training. 
More specifically, different from supervised learning, in which each image is used independently from other images, CL relies on diverse data to learn the correlation between different images.
Without large data diversity, the performance of CL performance will greatly degrade, which also results in low accuracy of the skin disease diagnosis after fine-tuning with labeled data.

To address this challenge, we propose a framework federated contrastive learning with feature sharing (FedCLF) to enable effective FL with limited labels. This framework has two stages. 
The first stage is federated self-supervised pre-training. Feature sharing is proposed to improve the data diversity of local contrastive learning while avoiding raw data sharing. Data features encoded in vectors are shared among devices, such that diverse and accurate contrastive information is provided to each device. 
By leveraging the shared features, representations of higher quality are learned on local devices, which improves the quality of the aggregated model in FL.

The second stage is fine-tuning with limited labeled data.
By using the pre-trained model in the first stage as a good initialization, the second stage learns the task of dermatological disease diagnosis by either fine-tuning independently on each device, or fine-tuning collaboratively on all devices by supervised federated learning with limited labels.

The \textbf{second} SSL category is generative self-supervised learning. 
The very recently developed method masked autoencoders (MAE) \cite{he2021masked} leverages masked autoencoding \cite{devlin2018bert} for learning from unlabeled data.
Different from CL which usually uses CNNs as the backbone, MAE uses Vision Transformers (ViT) as the backbone and leverages the sequence of image patches for SSL. During learning, MAE removes a portion of an input image and learns to predict the removed content. 
While the ViT-based MAE has demonstrated its advantages over CNN-based SSL methods in terms of learned representation quality and accuracy on downstream tasks in centralized training \cite{chen2021empirical, bao2021beit, he2021masked}, there is no existing work to investigate its performance in the FL setting with unlabeled data. 
Besides, the model synchronization of MAE between the server and distributed clients remains unexplored. In FL, usually, the whole model is synchronized \cite{mcmahan2017communication, li2020federated}. However, when applying MAE in FL, simply synchronizing the whole model may not be the best option.
More specifically, the MAE consists of an encoder and a decoder, both of which has a ViT-like architecture. 
When applied to the FL setting, the learned parameters of some components in the local MAE model can be tightly related to the local data and incorporate too much knowledge of the local data. Since the goal of FL is to learn a global and generalizable model with global knowledge, synchronizing all parameters will bring too much local knowledge to the global model and degrade its generalization performance.

To address these problems, we propose a framework FedMAE to systematically explore and improve the performance of MAE when combined with FL.
We first investigate the performance of FedMAE and demonstrate its superior accuracy in dermatological disease.
To further improve FedMAE, we explore the components in MAE to synchronize between the server and clients. Instead of synchronizing all the components, we split the MAE model into the local-knowledge components and global-knowledge components. Only the parameters of global-knowledge components are synchronized and the parameters of local-knowledge components are kept local without synchronization, which yields better representation quality and higher accuracy than synchronizing the entire model.

In summary, the main contributions of this paper include:
\begin{itemize}
	\item \textbf{\underline{Fed}erated \underline{c}ontrastive \underline{l}earning with \underline{f}eature sharing (FedCLF).} 
	We propose a FedCLF framework to enable effective learning with limited labels for dermatological disease diagnosis.
	FedCLF pre-trains the model on distributed unlabeled data to provide a good initialization, followed by fine-tuning with a limited number of labeled data to perform the disease diagnosis.
	\item \textbf{Feature sharing for better local learning.} 
	We propose a feature sharing method to improve the data diversity of local contrastive learning while avoiding raw data sharing for privacy.
	The shared features provide more diverse features to contrast with during local learning on each mobile device for better representations.
	\item \textbf{\underline{Fed}erated \underline{m}asked \underline{a}uto\underline{e}ncoder framework (FedMAE).} To further improve the accuracy of federated self-supervised learning, we propose FedMAE to leverage masked autoencoder (MAE) and ViT backbones to learn from decentralized data. We investigate the performance of FedMAE and demonstrate its superiority over CL with CNN backbones in federated self-supervised learning.
	\item \textbf{Global and local knowledge split and selective knowledge aggregation.} To learn a shared global model with good generalization performance, we split the knowledge learned by FedMAE into global knowledge and local knowledge. Only global knowledge is aggregated and local knowledge is kept local without interfering with the global model.
	\item \textbf{More accurate diagnosis and better label efficiency.} 
	Experiments on dermatological disease datasets show superior diagnostic accuracy and label efficiency over state-of-the-art techniques.
\end{itemize}

\section{Background and Related Work}

\begin{figure}[htb]
	\centering
	\includegraphics[width=0.75\linewidth]{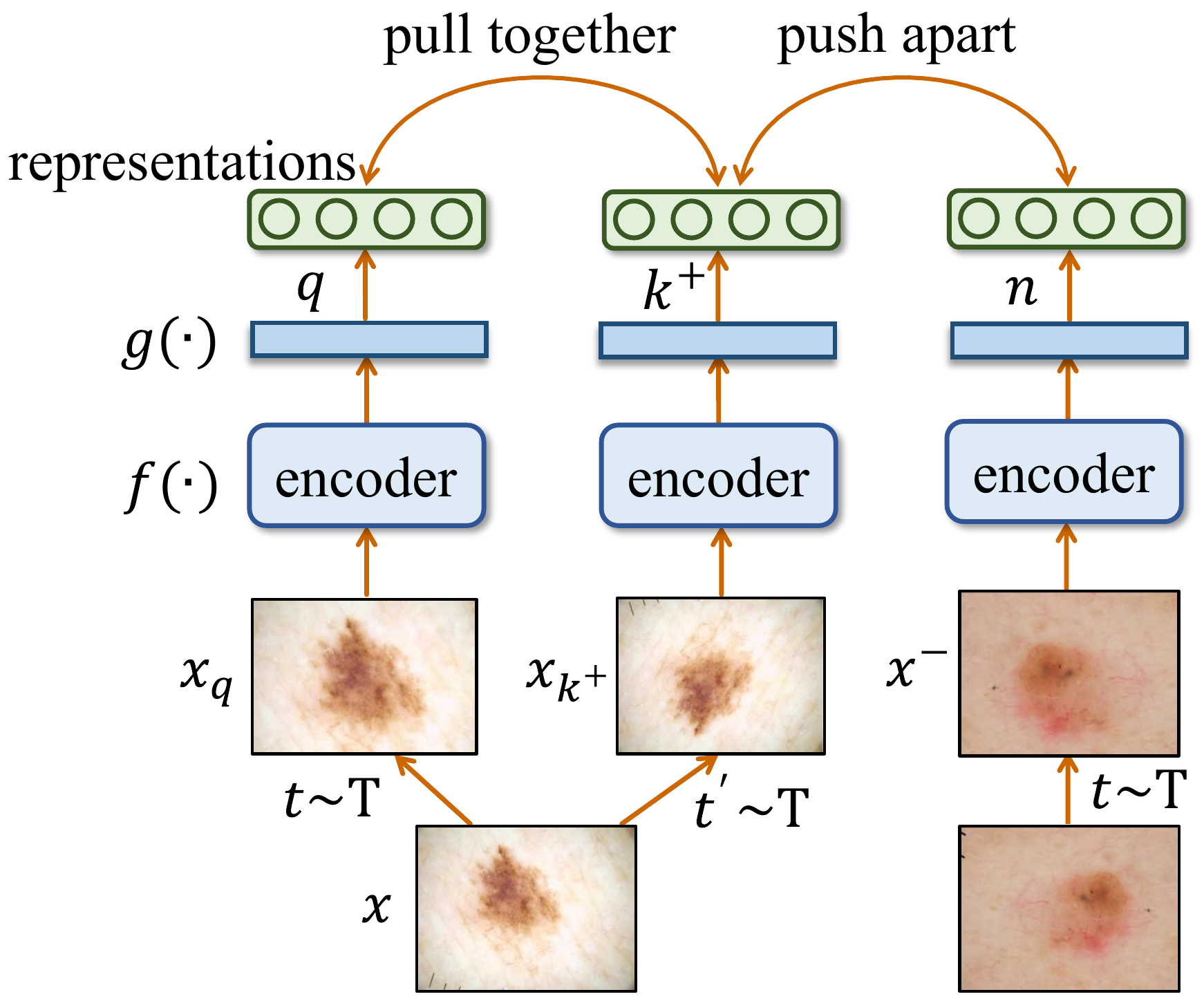}
	\caption{Illustration of contrastive learning. Two separate data augmentations operators are sampled from the same family of augmentations ($t\sim T$ and $t^{\prime} \sim T$) and then applied to one image $x$. The representations of the two transformed versions are pushing close to each other and apart from the representations of other images.}
	\label{fig:contrastive_learning_illustration}
\end{figure}

\begin{figure*}[ht]
	\centering
	\includegraphics[width=1.0\textwidth]{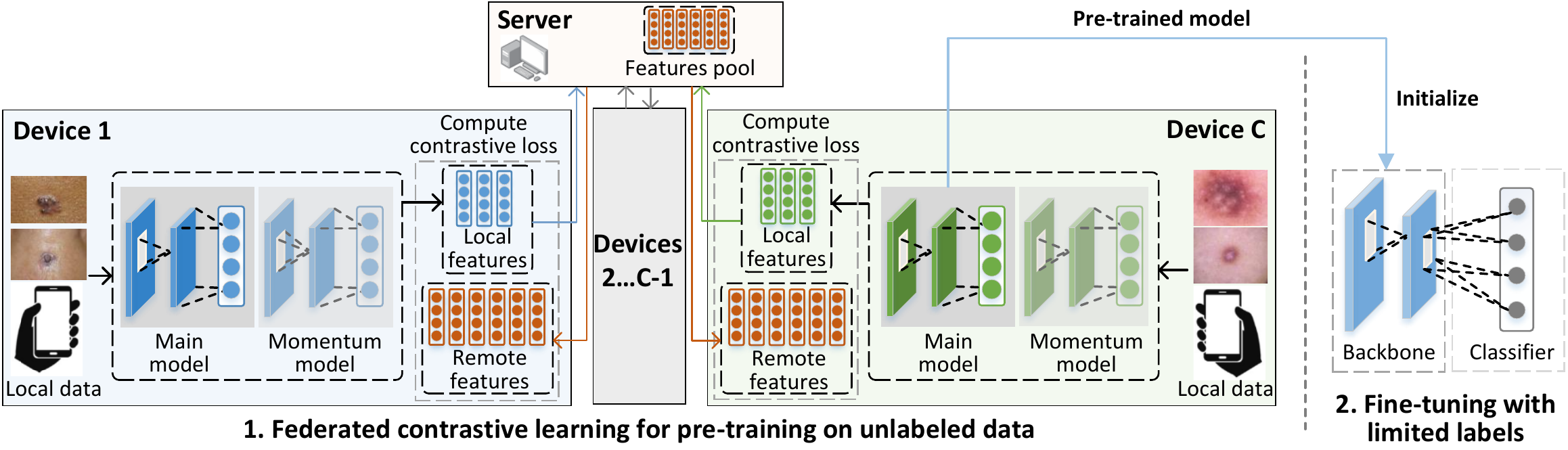}
	\caption{Federated contrastive learning with feature sharing (FedCLF) for pre-training the model with unlabeled data, by which good data representations are learned.
    After pre-training, the learned model is used as the initialization for fine-tuning with limited labels for dermatological disease classification.
	}
	\label{fig:overview}
\end{figure*}

\subsection{Contrastive Learning} 
Contrastive learning (CL) is a powerful self-supervised method to learn visual representations from unlabeled data \cite{tian2019contrastive,misra2020self,tian2020makes,wu2021enabling,chuang2020debiased}.
CL pre-trains a model and provides a high generalization performance for downstream tasks such as classification and segmentation \cite{he2020momentum,chen2020simple,chen2020big}.
CL learns representations by performing a proxy task of discriminating the image identities.
In the learning process, CL minimizes a contrastive loss evaluated on pairs of feature vectors extracted from data augmentations (e.g. cropping, rotation, and color distortion) of the image \cite{ho2020contrastive}.
By optimizing the contrastive loss, CL maximizes the agreement of representations between transformations of the same identity and minimizes the agreement between different images \cite{kim2020adversarial}.
As shown in Fig. \ref{fig:contrastive_learning_illustration}, for an unlabeled input image $x$, two random transformations $t \sim T$ and $t^{\prime} \sim T$ are applied to $x$ to produce $x_q$ and $x_{k^{+}}$, both of which are then fed into the model $f$ and projection head $g$ to get representations $q$ and $k^{+}$.
Let $Q$ be a memory bank with $K$ representation vectors stored. By using every feature $n$ in the memory bank $Q$ as negatives, a positive pair $q$ and $k^{+}$ are contrasted with every $n$ by the following contrastive loss function.
\begin{equation}\label{equ:mocoloss}
\ell_{q}=-\log \frac{\exp (q \cdot k^{+} / \tau)}{\exp (q \cdot k^{+} / \tau) + \sum_{n\in Q} \exp (q \cdot n / \tau)}.
\end{equation}
By optimizing the model $f$ to minimize the loss function, effective visual representations can be learned by $f$.

However, existing CL works are developed for centralized training on large-scale datasets consisting of millions \cite{imagenet_cvpr09} or even billions \cite{he2020momentum} of images.
The large-scale datasets provide sufficiently large data diversity for training.
However, in FL each device only has a limited amount of data with limited diversity. Since CL relies on the contrast with different data to achieve high performance, without large data diversity, the performance of CL will greatly degrade.

\subsection{Vision Transformers (ViT) and Masked Autoencoders (MAE)}

Transformer has become the de facto backbone for natural language processing (NLP) due to its long-range and self-attention mechanism \cite{vaswani2017attention, devlin2018bert}.
Recently, a lot of works apply Transformers for computer vision tasks \cite{dosovitskiy2021an,carion2020end,zhu2020deformable}.
Vision Transformers (ViT) \cite{dosovitskiy2021an} shows that pure Transformer-based architecture can also achieve SOTA results for image classification compared to CNN-based networks, especially for large-scale datasets.
To enable the superior performance of ViT on small datasets, DeiT \cite{touvron2021training} presents training strategies such as distillation token.
SwinTransformer \cite{liu2021swin} introduces the shifted window to greatly increase the efficiency of Transformer for dense vision tasks when the input image resolution is high such as object detection and semantic segmentation.

ViT has also been widely used for SSL problems.
For example, iGPT \cite{chen2020generative} masks and reconstructs image pixels, while MAE \cite{he2021masked} masks and reconstructs image patches based on ViT backbone.
IBOT \cite{zhou2022image} studies masked image modeling (MIM) and  performs self-distillation on masked patch tokens to acquire visual semantics for pre-training.
BEiT \cite{bao2021beit} tokenizes the original image into visual tokens by using discrete variational auto-encoder (DVAE) and proposes a MIM task to pre-train ViT.
Significant improvements can be achieved compared to other SSL baselines on the ImageNet-1k dataset.

However, existing ViT works focus only on centralized training. The performance of ViT and its SSL in the FL setting remains unexplored. 
In this work, we use MAE as our SSL framework because it performs better than other methods and does not require additional training such as the dVAE in BEiT, which makes it more suitable for federated learning.

\subsection{Federated Learning}
Federated learning (FL) aims to collaboratively learn a global model for distributed devices while keeping data on local devices for privacy \cite{mcmahan2017communication}.
In FL, the training data are distributed among devices, and each device has a subset of the training data.
In a typical FL algorithm FedAvg \cite{mcmahan2017communication}, 
learning is performed round-by-round by repeating the local learning and model aggregation process until convergence.
In one round, the server activates a subset of devices and sends them the latest model. Then the activated devices perform learning on local data. 
More specifically, in communication round $t$, the server activates a subset of devices $C^t$ and downloads the latest global model with parameters $\theta^t$ to them.
Each device $c\in C^t$ learns on private dataset $D_c$ by minimizing the local loss 
$\ell_c$ 
to get the updated local parameters $\theta_c^{t+1}$.
The locally updated models are aggregated into the global model by averaging the local parameters  
$\theta^{t+1} \leftarrow \sum_{c\in C^t}\frac{|D_c|}{\sum_{i\in C^t} |D_i|}{\theta_c^{t+1}}$.
This learning process continues until convergence.

The problem with these FL works is that they assume the devices have ground-truth labels for all the data. However, this assumption is not realistic in dermatological disease diagnosis due to the high labeling cost and the requirement of expertise for accurate labeling.
Therefore, to apply FL to dermatological disease diagnosis, one approach to effectively use limited labels to achieve high model accuracy is needed.

\begin{figure*}[t]
	\centering
	\includegraphics[width=0.8\textwidth]{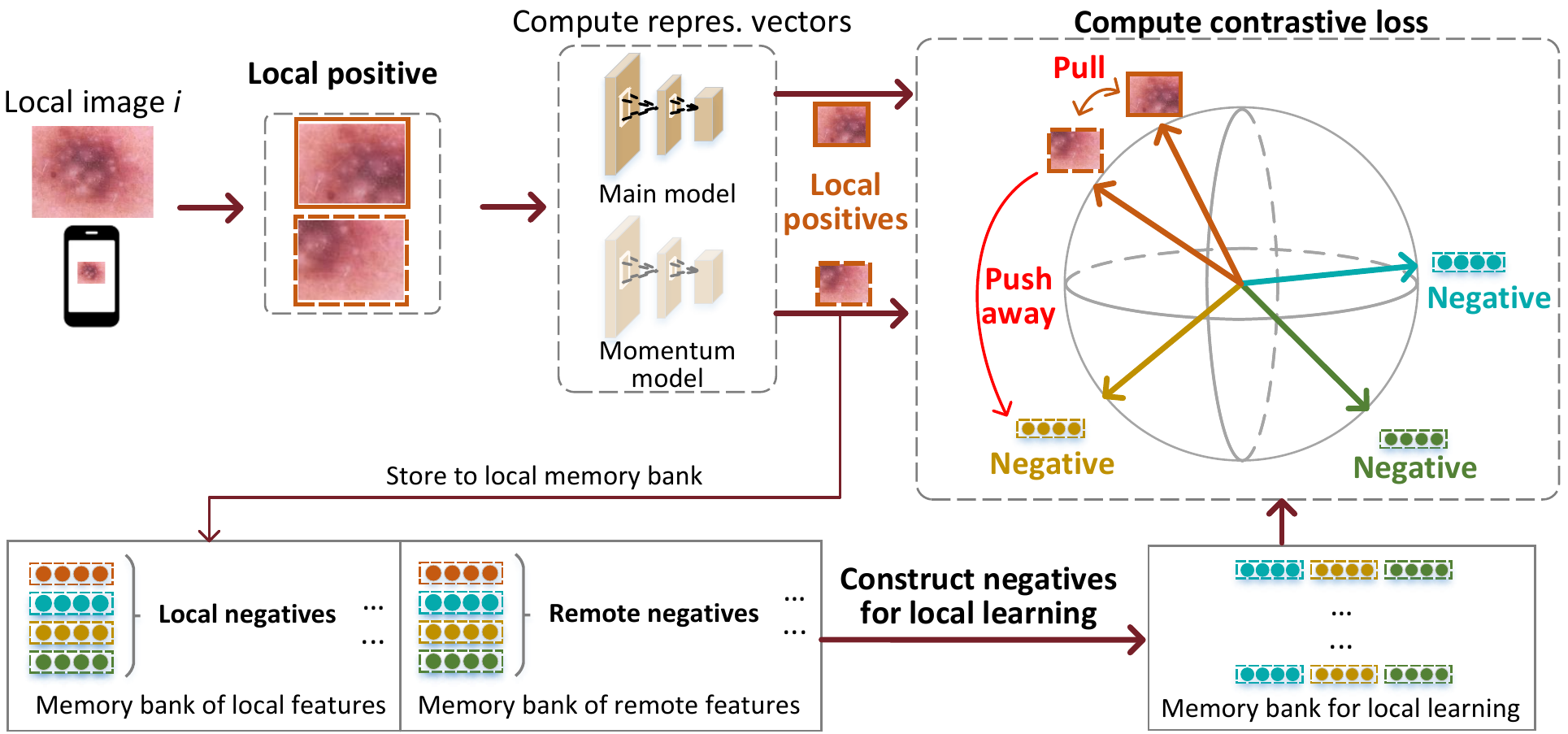}
	\caption{
    Local contrastive learning with shared features on one device in FedCLF. The remote features from other devices serve as negatives to improve the data diversity for better-learned representations. 
    During learning, the features of local positives are pushed close to each other and apart from the remote negatives.
	}
	\label{fig:feature_exchange}
\end{figure*}

Several concurrent works target federated pre-training on unlabeled data.
In \cite{wu2021federated, wu2022distributed, dong2021federated, wu2022decentralized, wu2021federated_skin}, encoded data representations or metadata are exchanged among clients for better local CL.
The proposed work differs from these works in that 
we employ Vision Transformers (ViT) as the backbone for federated self-supervised learning, which has demonstrated superior performance to convolutional neural networks (CNNs) in centralized training \cite{he2022masked} but remains unexplored in the federated setting.
Besides, we explicitly separate the local knowledge and global knowledge in the ViT model and only aggregate the global knowledge for better generalization performance.

\section{FedCLF: Federated Contrastive Learning with Remote and Local Features}\label{sec:fedclf}

In this section, we present our federated contrastive learning method. We will first describe the framework overview. 
Then we will present how to perform contrastive learning on each device with both local and remote features, by which a good initialization for fine-tuning can be learned.

\begin{figure*}[t]
	\centering
	\includegraphics[width=0.65\linewidth]{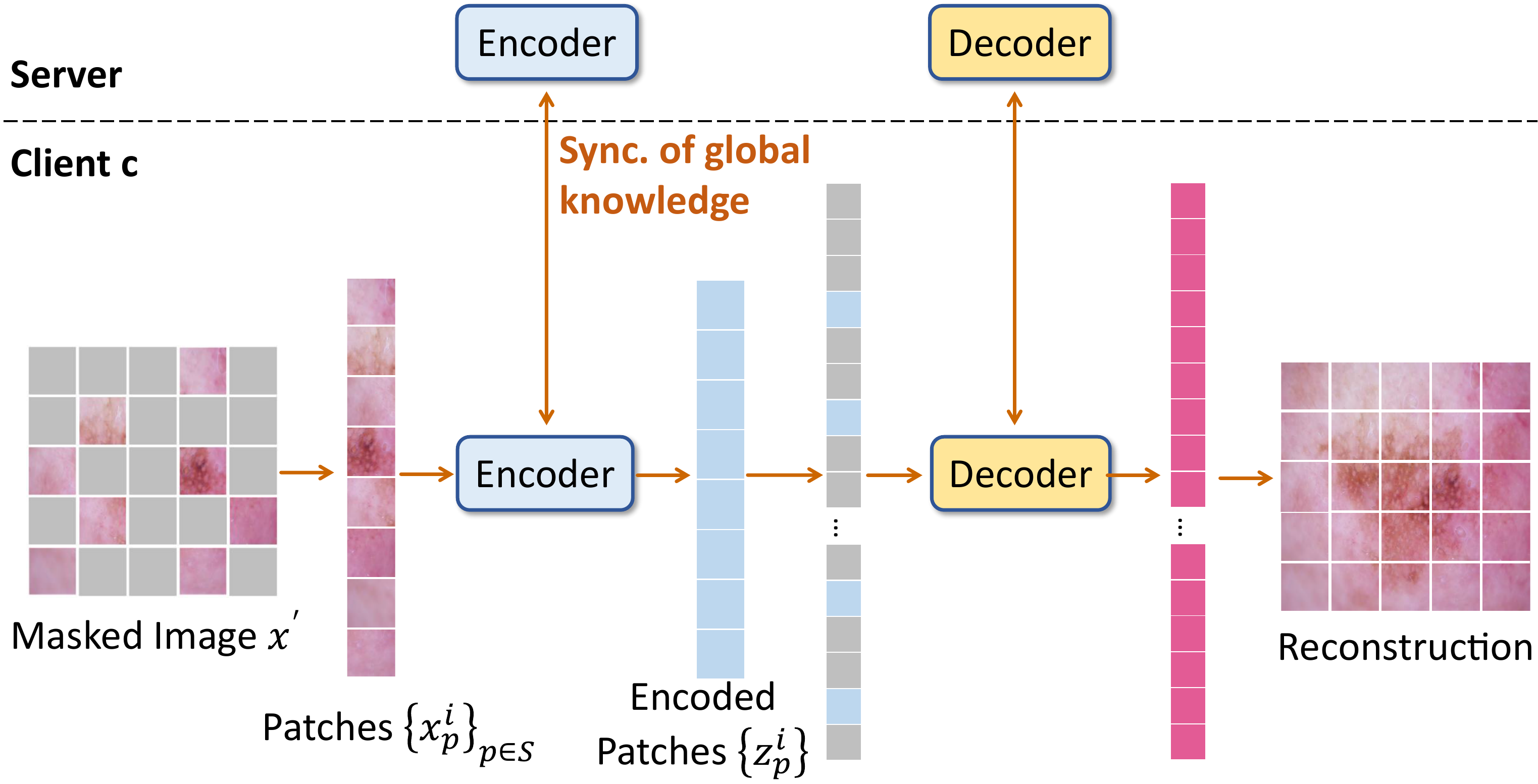}
	\caption{Illustration of federated masked autoencoders (FedMAE) based on ViT encoder and decoder.}
	\label{fig:fedmae_overview}
\end{figure*}

\subsection{Overview}

The overview of the proposed federated contrastive learning with feature sharing (FedCLF) framework is shown in Fig. \ref{fig:overview}. 
This framework has two stages. 
In the first stage, the model is collaboratively pre-trained by distributed devices on unlabeled data to extract visual representations. 

During learning, there is a server and many devices. The server is used to coordinate the learning process, aggregate the locally updated models and forward the shared features.
The devices perform CL on local data as well as local and remote features to update local models.
In the second stage, the model pre-trained in the first stage is used as the initialization for fine-tuning with limited labels. This can be achieved by existing supervised learning methods independently on each device, or collaboratively by supervised federated learning \cite{mcmahan2017communication}.
In this paper, we focus on the first stage of FedCLF on unlabeled data for pre-training.

To achieve better contrastive learning on each device, one can share raw data (i.e. skin images) with other devices to improve the data diversity for local learning. However, since skin images are highly sensitive and private, sharing them will cause serious privacy concerns for patients.
To improve data diversity while keeping raw data local for privacy, we proposed to share features (i.e. encoded vectors).
As shown in Fig. \ref{fig:overview}, FedCLF is performed round-by-round.
In one round, each device encodes its data as local features and uploads them to the server. Local models are also uploaded for model aggregation.
Then, the server de-identifies the uploaded features and sends these anonymous features to devices as remote features. 
The uploaded models are aggregated by averaging the weights of all models following \cite{mcmahan2017communication}, which serves as the initial model for the next round.
After that, the aggregated model and the remote features are distributed to devices.
The local models on devices are updated by using both local and remote features on each device, which provides more accurate and diverse contrastive information for computing the local contrastive loss.
Finally, the updated local models and features are uploaded to the server to initialize the next round.
This iterative learning process continues until the model convergence.

\subsection{Local Model Architecture}
The local contrastive learning process on a device is shown in Fig. \ref{fig:feature_exchange}.
We use the Momentum Contrast (MoCo) model architecture \cite{he2020momentum} since it uses a separate memory bank to store negative features, which can be filled with both local and remote negatives in the FedCLF setting.
Other popular contrastive learning models such as SimCLR \cite{chen2020simple} use features of data in the same mini-batch as negatives, which tightly couples the negative features with the mini-batch data and is not suitable to leverage remote features in FedCLF.
Different from SimCLR, MoCo decouples the negative features with the mini-batch data by a memory bank and can leverage the remote features as negatives for local contrastive learning.

On each device,
there are two models, the main model and the momentum model, and they have different functions.
The main model is the target model for learning and will be used as the initialization for the fine-tuning in the second stage.
The momentum model is used to generate features for local contrastive learning as local negatives and for sharing as remote features. 
The generated features will be stored in the memory bank of local features.
In the learning process, the main model is updated by the contrastive loss, and the momentum model is updated as the exponential moving average of the main model with a momentum coefficient.

\subsection{Memory Banks of Local and Remote Features}
To perform contrastive learning on each device,
the memory banks of local features and remote features need to be constructed such that they can provide features for computing the local loss. 
Each device has a memory bank of local features and a memory bank of remote features.
Local features are generated by feeding local images to the momentum model to produce the feature vectors. 
The feature vectors are then stored in the memory bank of local features.
On device $i$, let $Q_{l,i}$ be the memory bank of local features with $K$ features stored. $Q_{l,i}$ is used as local negatives and maintained following a first-in-first-out principle. The oldest features in the local memory bank will be dropped when new features are generated.

The remote features are collected from other clients.
At the beginning of each round of FedCLF, the local features are uploaded to the server and the remote features will be downloaded to fill the memory bank of remote features. More specially, the memory bank of remote features $Q_{r,i}$ on device $i$ is filled as follows.
\begin{equation}\label{equ:remote_neg}
Q_{r,i} = \{Q_{l,c}\ |\ 1 \le c \le |C|, c\ne i \},
\end{equation}
where $C$ is the set of all devices.

\subsection{Learning with Remote Features}

\noindent
\textbf{Constructing negatives from memory banks for local learning.}
By leveraging memory banks of local features $Q_{l,i}$ and remote features $Q_{r,i}$, the negatives $Q_{\text{CL},i}$ for computing the contrastive loss is constructed as follows. 
For conciseness, we leave out the device index $i$ in $Q_{\text{CL},i}$.

On device $i$,
at the beginning of each round of FedCLF, $Q_{\text{CL}}$ is initialized as the memory bank of local features $Q_{l,i}$. 
During each mini-batch of local learning, a mini-batch data $x$ of size $B$ is fed into the momentum model to generate the features $q_{l,i,B}$. Then $B$ remote features are sampled uniformly from remote features $Q_{r,i}$ as follows.
\begin{equation}\label{equ:contrastive_negatives}
    q_{r,i,B}=\{ Q_{r,i,j} |\ j \sim \mathcal{U}(|Q_{r,i}|,B) \},
\end{equation}
where $j \sim \mathcal{U}(|Q_{r,i}|,B)$ means 
$B$ indices are sampled uniformly from the range $[0,|Q_{r,i}|-1]$ and $Q_{r,i,j}$ is the $j$-$th$ feature in $Q_{r,i}$.

After learning a mini-batch, the oldest features in $Q_{\text{CL}}$ will be replaced by the latest features $q_{\text{update}}$ as follows.
\begin{equation}\label{equ:q_update}
q_{\text{update}} = \{q_{l,i,B} \cup q_{r,i,B}\}.
\end{equation}

\noindent
\textbf{Removing local negatives for more accurate learning.}
While $Q_{\text{CL}}$ updated by Eq.(\ref{equ:q_update}) contains remote features for improved data diversity, $Q_{\text{CL}}$ also contains local features. 
For better local learning, we propose to completely avoid using local features as negatives during local learning. 
The intuition is that local features can share certain levels of similarity with the data that is being learned because they are from the same patient.
Using local features as negatives can degrade the learned representations since it pushes the representations of the data being learned apart from the local features, which could have been clustered for better representations.

To solve this problem, we eliminate the use of local features during local contrastive learning. At the beginning of each round of FedCLF, $Q_{\text{CL}}$ is initialized as the memory bank of remote features $Q_{r,i}$ in Eq.(\ref{equ:remote_neg}) instead of $Q_{l,i}$.
The latest features for updating $Q_{\text{CL}}$ are also simplified as follows.
\begin{equation}\label{equ:q_update_no_local_neg}
q_{\text{update}} = \{q_{r,i,B}\}.
\end{equation}

Compared with Eq.(\ref{equ:q_update}), the local negatives are removed in Eq.(\ref{equ:q_update_no_local_neg}).
In this way, better representations can be learned during FedCLF, which also results in a higher accuracy of the diagnostic model after fine-tuning.

\noindent
\textbf{Loss function.}
On device $i$, by using the constructed $Q_{\text{CL}}$, the feature $q$ of one image being learned is compared with all features in $Q_{\text{CL}}$, and the contrastive loss for $q$ is defined as:
\begin{equation}\label{equ:loss_memorybank}
\ell_{q,k^{+},Q_{\text{CL}}} = - \log \frac{\exp (q \cdot k^{+} / \tau)}{\exp (q \cdot k^{+} / \tau) + \sum_{n\in Q_{\text{CL}}} \exp (q \cdot n / \tau)},
\end{equation}
where the operator $\cdot$ is the dot product between two vectors and $\tau$ is the temperature to control the distribution concentration degree \cite{hinton2015distilling}.
By minimizing the loss, the representations of local data can be effectively learned.

\section{FedMAE: Federated Masked Autoencoders}

In Section \ref{sec:fedclf}, we have proposed FedCLF for federated self-supervised learning with CNNs as the backbone. 
However, the performance of the self-supervised Vision Transformer (ViT) in the FL setting has not been explored. 
In this section, we present our federated self-supervised learning with masked autoencoders (FedMAE) to enable federated self-supervised learning with ViT.
We first revisit the ViTs. Then we describe the overview of our FedMAE for dermatological disease diagnosis.
After that, we introduce the local training with unlabeled data on each client.
Finally, we propose the global and local knowledge split for better knowledge aggregation.

\subsection{Revisiting Vision Transformers (ViT)}

ViTs are based on the NLP Transformer architecture BERT \cite{devlin2018bert} with adaptation layers for image inputs and task outputs.
Transformer uses alternating layers of multiheaded self-attention (MSA) and multi-layer perception (MLP) blocks as the backbone \cite{dosovitskiy2021an}. 
Since the standard Transformer model takes 1D sequence of token embeddings as input, to handle 2D images, an input image $x \in \mathbb{R}^{H \times W \times C}$ is reshaped into $N=HW/P^2$ patches $x_p \in \mathbb{R}^{N \times (P^2 \cdot C)}$, where $(H, W)$ is the input image resolution, $C$ is the number of channels, and $(P,P)$ is the patch resolution.
The image patches $\{x_p^i\}_{i=1}^{N}$ are flattened into vectors and linearly projected as patch embeddings $\{z_i\}_{i=1}^{N}$. 
Then, a learnable class embedding ($z_0=x_\text{class}$) is appended to the patch embeddings, whose state at the Transformer output is the visual representation of the input image.

\subsection{Framework Overview}

The overview of the proposed FedMAE is shown in Fig. \ref{fig:fedmae_overview}. There is a server and multiple clients. Each client trains its local MAE model \cite{he2021masked} by performing an image reconstruction task to learn visual representations with unlabeled data. 
The learning is performed round-by-round. 
At the beginning of each round, the MAE model is distributed from the server to clients, after which local training begins on clients.
During local training, given an unlabeled image, a large random subset of image patches is masked out. The remaining visible patches are fed into the encoder. The encoded patches together with mask tokens are then fed into the decoder for pixel-level reconstruction of the original full image. 
In the end of each round, the MAE model is aggregated from clients to the server following the model averaging scheme FedAvg \cite{mcmahan2017communication}.
After learning, the encoder is used as the initialization for fine-tuning with limited labels.

By leveraging the powerful backbone ViT \cite{dosovitskiy2021an} and self-supervised training method MAE \cite{he2021masked}, our seemingly naive, yet surprisingly effective FedMAE can learn better representation and achieve higher accuracy for dermatological disease diagnosis than CL-based methods which usually use CNNs.

\subsection{Local Training}
During local training, each client performs learning on unlabeled data by using MAE \cite{he2021masked}. There is an encoder $f_\theta$ that generates representations by taking an image as input, and a decoder $g_\theta$ that reconstructs the original image from the (partial) representations. Both the encoder and the decoder follow the ViT architecture. The encoder will be used for dermatological disease diagnosis and the decoder will be discarded.

As shown in Fig. \ref{fig:fedmae_overview}, local learning starts by taking an input image $x$ as input and dividing it into non-overlapping patches. Then a subset of the patches is sampled uniformly at random as the masked image $x^\prime = \{x_p^i\}_{i\in S}$ and other patches are removed, where $S$ denotes the indices of selected patches.
The selected patches $\{x_p^i\}_{i\in S}$ are fed into the encoder to generate encoded patches (representations) $\{z_p^i\}_{i\in S}$. 

The input to the decoder consists of the tokens for the full-sized image, where the tokens for the selected patches are $\{z_p^i\}_{i\in S}$ generated by the encoder, while the tokens for the removed (masked) patches are the mask token $z_m$. 
The mask token $z_m$ is a learnable vector to indicate the missing patches to be reconstructed. The output of the encoder is a set of vectors, each representing the pixel values of a patch.

The learning is performed by pixel-level reconstruction of the original image and updating parameters $\theta$ with the reconstruction loss:
\begin{equation}
    \mathcal{L}_{\theta} = \norm{x - g_\theta(f_\theta(x^\prime))} ^2_2 ,
\label{eq:mae_loss}
\end{equation}
where $g_\theta(f_\theta(x^\prime))$ is the reconstructed image by the decoder, and $x$ is the original image.

In each round, the local MAE model is trained for $E$ epochs before uploading the model to the server for aggregation. 

\subsection{Global and Local Knowledge Split}\label{sect:knowledge_split}

\begin{figure}[htb]
	\centering
	\includegraphics[width=0.9\linewidth]{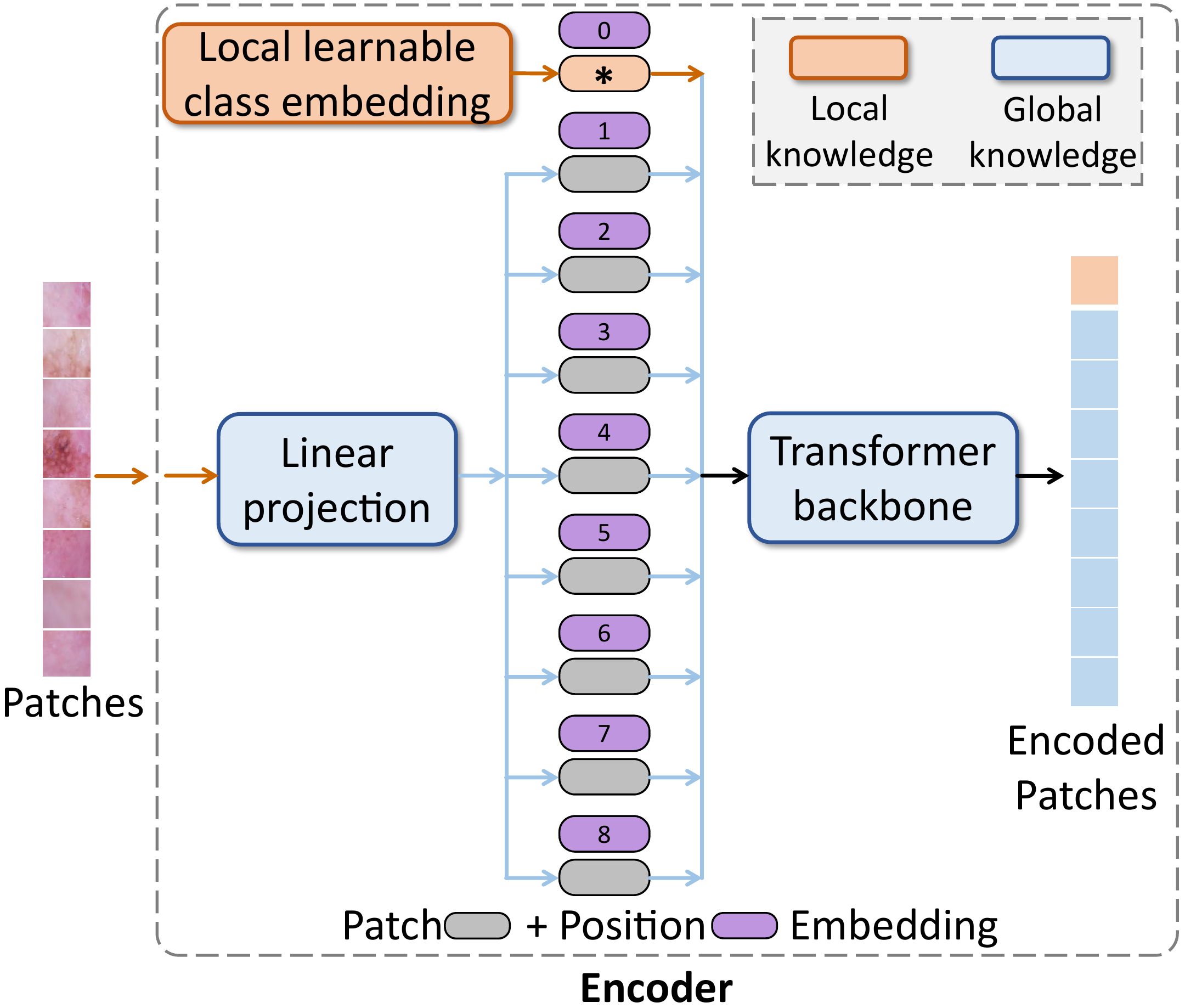}
	\caption{Illustration of global and local knowledge of FedMAE during learning on a client. The knowledge of the encoder in MAE is split into global knowledge and local knowledge. Only the components containing global knowledge (denoted in blue color) are synchronized with the server, while the component containing local knowledge (i.e. the learnable class embedding denoted in orange color) is not synchronized.}
	\label{fig:fedmae_global_local}
\end{figure}

Following the standard FL model aggregation method FedAvg \cite{mcmahan2017communication}, it is straightforward to aggregate all the parameters from the local MAE models for FedMAE. 
However, since FedAvg is originally designed for FL with CNNs and does not consider the characteristics of ViT architecture in FedMAE, synchronizing all the parameters may not be the best option. 

To address these problems and learn a global model with a high generalization performance, it is beneficial to split the learned knowledge of the MAE model into global knowledge and local knowledge. Only aggregating the parameters containing global knowledge while not synchronizing those parameters with too much local and divergent information will improve the generalization performance of the aggregated model.
Following this intuition and thanks to the highly modular architecture of ViT in MAE, 
we empirically observe that the knowledge learned in the encoder of MAE can be naturally split into global and local knowledge. As shown in Fig. \ref{fig:fedmae_global_local}, the component in orange color (i.e. the learnable class embedding) highly depends on the local data it is trained on and is treated as local knowledge. Other components in blue color, including the linear projection layer and the Transformer backbone, contains global knowledge beneficial to the global model and are treated as global knowledge. 

Based on the knowledge split, during FL rounds, the components containing global knowledge (i.e. all components in the MAE except for the learnable class embedding in the encoder) are synchronized following FedAvg. The learnable class embedding is kept local and only aggregated once at the end of the last round of the FedMAE learning process.

\section{Experiments}

In this section, we first present the experimental results of FedCLF in Section \ref{sec:res_fcl}. Then we describe the results of FedMAE in Section \ref{sec:res_fedmae}.
FedCLF and FedMAE have different computation costs and accuracy, where FedMAE features high accuracy with $3.5\times$ training time of FedCLF.

\subsection{Results of FedCLF}\label{sec:res_fcl}

\noindent
\textbf{Datasets.}
The proposed FedCLF method is evaluated on four datasets of different skin colors, including the ISIC 2019 challenge dataset \cite{ISIC2019} mainly for white skins, AtlasDerm \cite{AtlasDerm}, and Dermnet \cite{Dermnet} mainly for brown skins, and DarkDerm mainly for dark skins collected by us.
Since these four datasets have a different number of diagnostic categories, to form a unified classification task, we use their intersection of five diseases, including basal cell carcinoma (BCC), dermatofibroma (DF), melanoma (MEL), melanocytic nevus (NV), and squamous cell carcinoma (SCC). 
ISIC dataset consists of about 25k dermoscopic images among nine different diagnostic categories, and we use a subset with about 21k images in the above five classes.
AtlasDerm has about 11k images in 560 categories, and we use its subset in the above five classes with 618 images.
Dermnet consists of images in 23 types of dermatology diseases, and we use a subset in the above five classes with 276 images.
DarkDerm is established with 216 images in the above five categories.
In the pre-processing, the images are resized with bi-linear interpolation such that the dimension of the shorter edge is 72 pixels while keeping the original aspect ratio.

\noindent
\textbf{Federated setting.}
We use 10 devices for FL and distribute datasets based on skin colors to simulate different patients.
The ISIC dataset is randomly split into 7 partitions and each partition is distributed to one of the first 7 devices.
The AtlasDerm, Dermnet, and DarkDerm datasets are assigned to one of the following three devices, respectively.
On each device, the assigned dataset is randomly split into a training set, a validation set and a test with 60\%, 20\% and 20\% data, respectively. 
We use ResNet-18 \cite{he2016deep} as the backbone model, and use a 2-layer MLP projection head to project the representations to 128-dimensional features.

\noindent
\textbf{Evaluation.}
We use the proposed FedCLF method to pre-train the model by the distributed devices without using labels.
Then the pre-trained model is used as the initialization for fine-tuning with limited labels.
We consider two practical settings for fine-tuning, \textit{local fine-tuning} and \textit{federated fine-tuning}. 
In local fine-tuning, each device independently fine-tunes its model with its limited labeled data after pre-training.
In federated fine-tuning, devices collaboratively fine-tune the pre-trained model with limited labeled data by supervised federated learning.
During fine-tuning, we evaluate with different fractions of labels, where the percentage of labeled data in the training set is $L\in \{10\%,20\%,40\%,80\%\}$ on each device.
Following \cite{wu2019studies}, we use two metrics for evaluation, including the mean recall of each class (i.e. balanced multiclass accuracy used as the primary metric for the ISIC 2019 challenge \cite{ISIC2019}) and the mean precision of each class.
We report the mean recall and mean precision on the test set of all devices.

\noindent
\textbf{Training details.}
The pre-training of FedCLF is performed for 100 communication rounds, and FedAvg \cite{mcmahan2017communication} is employed as the model aggregation algorithm on the server in each round.
The active device ratio per round is 1.0 and the number of local training epochs before each aggregation is 1.
The batch size is 128 and the initial learning rate is 0.03 with a cosine decay schedule.
In the fine-tuning stage, the model is trained for 20 epochs in local fine-tuning or 100 rounds in federated fine-tuning.
In local fine-tuning, the Adam optimizer is used with a batch size of 256, a learning rate of 1e-4 with a decay factor of 0.2 at epochs 12 and 16.
In federated fine-tuning, the batch size is 128 and the learning rate is 1e-4.
The training is performed on one Nvidia 2080Ti GPU.

\noindent
\textbf{Baselines.}
We compare the proposed techniques with four baselines for pre-training.
\textit{Random init} uses random model initialization for fine-tuning.
\textit{Local CL} pre-trains the model by contrastive learning independently on each device with unlabeled data.
\textit{Rotation} \cite{gidaris2018unsupervised} is a self-supervised learning approach for pre-training by predicting the rotation angles of images.
\textit{SimCLR} \cite{chen2020simple} is a SOTA contrastive learning based approach.
We combine these two methods with the FL framework FedAvg \cite{mcmahan2017communication} as \textit{FedRotation} and \textit{FedSimCLR}.

\subsubsection{Local Fine-tuning}

\begin{table}[!htb]
	\centering
	\caption{Results of \textbf{local fine-tuning} by the proposed FedCLF and baselines. The model is pre-trained by different approaches without using labels and then fine-tuned with limited labeled data independently on each device. $L$ is the label fraction on each device for fine-tuning. 
	The recall and precision averaged over all devices are reported, and on each device, the recall and precision are averaged over all classes.
	With different label fractions, consistent improvements by the proposed approaches over the baselines are observed.
	}
	\label{tab:exp_local_finetune}
	\setlength\tabcolsep{2.0pt}
	\renewcommand{\arraystretch}{1.0}
	\resizebox{1.0\linewidth}{!}{
    \begin{tabular}{lcccccccc}
    \toprule
    \multirow{2}{*}{} & \multicolumn{2}{c}{$L$=10\%} & \multicolumn{2}{c}{$L$=20\%} & \multicolumn{2}{c}{$L$=40\%} & \multicolumn{2}{c}{$L$=80\%} \\
    Methods     & Recall & Prec. & Recall & Prec. & Recall & Prec. & Recall & Prec.  \\ \midrule
    Random Init. & 21.56 & 17.35 & 23.13 & 20.79 & 24.88 & 23.69 & 23.92 & 26.06 \\
    Local CL \cite{chaitanya2020contrastive}   & 26.57 & 26.54 & 28.82 & 28.20 & 31.46 & 30.49 & 34.27 & 30.71 \\
    FedRotation \cite{gidaris2018unsupervised} & 23.45 & 25.27 & 25.05 & 25.05 & 29.39 & 28.48 & 32.87 & 28.15 \\
    FedSimCLR \cite{chen2020simple}  & 30.11 & 31.25 & 32.77 & 31.67 & 35.50 & 33.09 & 37.58 & 33.85 \\
    FedCLF (ours)    & \textbf{30.41} & \textbf{33.54} & \textbf{34.41} & \textbf{32.96} & \textbf{37.03} & \textbf{34.95} & \textbf{39.25} & \textbf{35.02} \\ \bottomrule
    \end{tabular}
    }
\end{table}

We compare the performance of different methods by local fine-tuning with limited labels. The results are shown in Table \ref{tab:exp_local_finetune}, and both recall and precision are reported.
The proposed methods effectively improve the recall and precision with different fractions of labeled data.
First, with 10\%, 20\%, 40\% and 80\% labeled data on each device for fine-tuning, the proposed approaches outperform the best-performing baseline by 0.30\%, 1.64\%, 1.53\%, 1.67\% for recall, and 2.29\%, 1.29\%, 1.86\%, 1.17\% for precision, respectively.
Second, the proposed approaches effectively use limited labels for fine-tuning. 
With 40\% labels, the proposed approaches achieve 37.03\% recall and 34.95\% precision, which are on par with or even better than the best-performing baseline with $2\times$ labels (37.58\% recall and 33.85\% precision).

\begin{table}[!htb]
	\centering
	\caption{
	Results of \textbf{federated fine-tuning} by the proposed FedCLF and baselines.
	The model is pre-trained by different approaches without using labels and then fine-tuned with limited labeled data collaboratively by all devices. 
	$L$ is the label fraction for fine-tuning on each device. 
	The recall and precision averaged over all classes are reported.
	With different label fractions, consistent improvements by the proposed approaches over the baselines are observed.
	}
	\label{tab:exp_federated_finetune}
	\setlength\tabcolsep{2.0pt}
	\renewcommand{\arraystretch}{1.0}
	\resizebox{1.0\linewidth}{!}{
    \begin{tabular}{lcccccccc}
    \toprule
    \multirow{2}{*}{} & \multicolumn{2}{c}{$L$=10\%} & \multicolumn{2}{c}{$L$=20\%} & \multicolumn{2}{c}{$L$=40\%} & \multicolumn{2}{c}{$L$=80\%} \\
    Methods     & Recall & Prec. & Recall & Prec. & Recall & Prec. & Recall & Prec.  \\ \midrule
    Random Init. & 43.15 & 38.97 & 45.63 & 40.41 & 50.73 & 44.66 & 55.61 & 46.35 \\
    Local CL \cite{chaitanya2020contrastive}   & 43.41 & 39.59 & 45.69 & 41.39 & 50.35 & 45.58 & 55.47 & 47.70 \\
    FedRotation \cite{gidaris2018unsupervised} & 43.18 & 39.57 & 44.36 & 40.27 & 50.69 & 44.10 & 54.27 & 46.70 \\
    FedSimCLR \cite{chen2020simple}  & 45.89 & 41.44 & 48.92 & 43.39 & 54.17 & 46.71 & 58.64 & 48.27 \\
    FedCLF (ours)    & \textbf{48.03} & \textbf{42.87} & \textbf{51.50} & \textbf{45.71} & \textbf{55.13} & \textbf{48.73} & \textbf{59.23} & \textbf{50.21} \\ \bottomrule
    \end{tabular}
    }
\end{table}

\subsubsection{Federated Fine-tuning}

We compare different methods by federated fine-tuning on all devices with limited labels. The results are shown in Table \ref{tab:exp_federated_finetune}, and both recall and precision are reported.
First, the proposed methods outperform the baselines by a large margin with different fractions of labeled data.
With 10\%, 20\%, 40\% and 80\% labeled data on each device for collaborative fine-tuning, the proposed approaches outperform the best-performing baseline by 2.14\%, 2.58\%, 0.96\%, 0.59\% for recall, and 1.43\%, 2.32\%, 2.02\%, 1.94\% for precision, respectively.
Second, the proposed approaches effectively improve the labeling efficiency. 
For instance, with 10\% labels, the proposed approaches achieve a similar performance as the best-performing baseline with $2\times$ labels (48.03\% vs. 48.92\% for recall and 42.87\% vs. 43.39\% for precision).

\subsubsection{Ablation Study}

\begin{figure}[!htb]
	\centering
	\includegraphics[width=0.7\columnwidth]{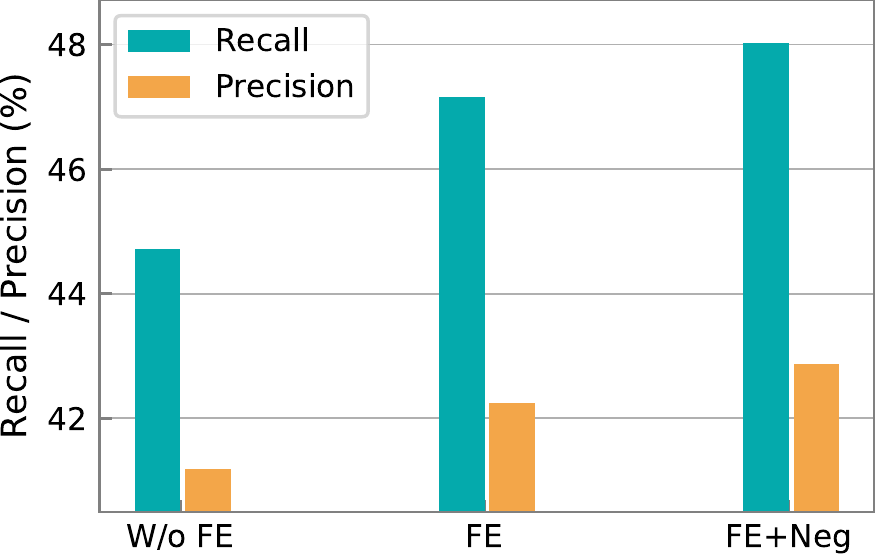}
	\caption{Ablation study of FedCLF. W/o FE is the naive approach without the proposed feature exchange, and FE is the approach with feature exchange enabled. FE+Neg further removes the local negatives. Results by federated fine-tuning with 10\% labeled data are reported. Each of the proposed approaches effectively improves recall and precision.}
	\label{fig:exp_ablation}
\end{figure}

We perform an ablation study to evaluate the effectiveness of each of the proposed approaches in FedCLF. We evaluate approaches by federated fine-tuning with 10\% labels, and the results are shown in Fig. \ref{fig:exp_ablation}.
For example, without feature exchange, the recall is 44.72\%. By enabling feature exchange, the recall is improved to 47.45\%. By further removing local negatives, the recall is improved to 48.03\%.
This result shows that each of the proposed approaches effectively improves the learned representations and improves the recall and precision of the dermatological disease diagnosis.

\subsection{Results of FedMAE}\label{sec:res_fedmae}
We follow most of the experimental setup in Section \ref{sec:res_fcl} and describe the additional setup for FedMAE as follows.

\noindent
\textbf{Datasets and federated setting.}
We evaluate the proposed FedMAE on the full ISIC 2019 challenge dataset \cite{ISIC2019} dataset with about 25k dermoscopic images in eight different diagnostic categories.
We randomly split the dataset into three partitions, with 60\% for training, 20\% for validation, and 20\% for test. 
In the pre-processing, the images are resized to 224$\times$224.
Following \cite{he2021masked}, we use data augmentations including random resized cropping and random horizontal flipping.
We randomly split the training set into 10 partitions and each of the 10 clients is assigned one partition.
The best global model is selected on the validation set. We report the test performance regarding multiple metrics including recall(sensitivity), precision, F1-score, specificity, AUC, and accuracy. All results are reported with an average of 3 independent runs.

\noindent
\textbf{Training details.}
We follow the training setup described in MAE \cite{he2021masked}.
We use ViT-Base (ViT-B) \cite{dosovitskiy2021an} as the backbone model for MAE. 
The image masking ratio is 75\%. The number of local epochs between each aggregation is 10.
The AdamW optimizer is used with a batch size of 256 and a learning rate of 1.5e-4 with a cosine decay schedule.
In the fine-tuning stage, the model is trained for 200 epochs in local fine-tuning and 50 rounds in federated fine-tuning.
In local fine-tuning, the batch size is 512 and the learning rate is 1e-3 with a cosine decay schedule.
In federated fine-tuning, the batch size is 256 and the learning rate is 5e-4.
For the CNN-based baseline methods, following \cite{zhou2022image,chen2021empirical}, ResNet-50 is used as the backbone.
The training is performed on one or two Nvidia V100 GPUs.

\subsubsection{Local Fine-tuning}

\begin{table}[!htb]
	\centering
	\caption{
	Results of \textbf{local fine-tuning} by FedMAE and baselines. 80\% data is labeled on each device for fine-tuning. 
	}
	\label{tab:exp_fedmae_local_finetune}
	\setlength\tabcolsep{2.0pt}
	\renewcommand{\arraystretch}{1.0}
	\resizebox{1.0\linewidth}{!}{
    \begin{tabular}{lcccccccc}
    \toprule
    Methods     & Recall & Precision & F1 & Specificity & AUC & Accuracy  \\ \midrule
    \multicolumn{7}{c}{\textit{CNN-based}} \\
    Random Init. & 32.65 & 36.23 & 33.93 & 92.45 & 74.64 & 58.92 \\
    Local CL \cite{chaitanya2020contrastive}   & 32.93 & 40.87 & 34.99 & 92.89 & 75.70 & 62.51 \\
    FedRotation \cite{gidaris2018unsupervised} & 33.38 & 33.49 & 32.49 & 92.45 & 74.26 & 59.47 \\
    FedSimCLR \cite{chen2020simple}  & 35.12 & 40.85 & 37.09 & 93.01 & 77.07 & 62.00 \\
    FedMoCo \cite{he2020momentum}  & 34.36 & 40.43 & 36.31 & 93.02 & 76.93 & 62.30 \\
    FedCLF (ours)   & 37.52 & 44.23 & 39.80 & 93.22 & 77.35 & 63.38 \\ \midrule
    \multicolumn{7}{c}{\textit{ViT-based}} \\
    Random Init.    & 35.69 & 38.15 & 36.56 & 92.69 & 77.42 & 59.70 \\ 
    FedMAE (ours)    & \textbf{41.31} & \textbf{45.41} & \textbf{42.76} & \textbf{93.63} & \textbf{80.71} & \textbf{64.80}  \\ \bottomrule
    \end{tabular}
    }
\end{table}

We evaluate the proposed methods by fine-tuning independently on each client, with 80\% local data labeled for fine-tuning. The model is fine-tuned on top of the encoder learned with unlabeled data by different approaches. 
The results are shown in Table \ref{tab:exp_fedmae_local_finetune}.
\textit{First}, our FedMAE significantly outperforms all the CNN-based methods. 
For example, our FedMAE outperforms the best-performing CNN-based baseline FedSimCLR by $+5.67\%$ F1-score and $+3.64\%$ AUC, respectively.
\textit{Second}, our FedMAE outperforms randomly initialized ViT, which verifies the effectiveness of self-supervised representation learning by FedMAE from unlabeled data.
\textit{Third}, our ViT-based FedMAE achieves better performance than our CNN-based FedCLF, showing the promise of using ViT backbones for federated self-supervised learning instead of CNN backbones.

\subsubsection{Federated Fine-tuning}

\begin{table}[!htb]
	\centering
	\caption{
	Results of \textbf{federated fine-tuning} by FedMAE and baselines. 80\% data is labeled on each device for fine-tuning. 
	}
	\label{tab:exp_fedmae_federated_finetune}
	\setlength\tabcolsep{2.0pt}
	\renewcommand{\arraystretch}{1.0}
	\resizebox{1.0\linewidth}{!}{
    \begin{tabular}{lcccccccc}
    \toprule
    Methods     & Recall & Precision & F1 & Specificity & AUC & Accuracy  \\ \midrule
    \multicolumn{7}{c}{\textit{CNN-based}} \\
    Random Init. & 33.69 & 57.34 & 35.17 & 92.63 & 82.19 & 64.87 \\
    Local CL \cite{chaitanya2020contrastive}   & 40.98 & 59.24 & 45.26 & 93.93 & 86.33 & 69.32  \\
    FedRotation \cite{gidaris2018unsupervised} & 34.35 & 46.00 & 35.46 & 93.42 & 83.13 & 65.92 \\
    FedSimCLR \cite{chen2020simple}  & 45.70 & 59.10 & 47.90 & 94.21 & 87.85 & 70.30 \\
    FedMoCo \cite{he2020momentum}  & 43.34 & 58.94 & 47.07 & 94.13 & 87.25 & 70.30 \\
    FedCLF (ours)   & 46.64 & 60.67 & 49.17 & 94.20 & 88.27 & 70.72 \\ \midrule
    \multicolumn{7}{c}{\textit{ViT-based}} \\
    Random Init.    & 51.55 & 57.79 & 54.18 & 94.56 & 86.82 & 70.30 \\ 
    FedMAE (ours)    & \textbf{58.08} & \textbf{65.76} & \textbf{61.09} & \textbf{95.32} & \textbf{90.26} & \textbf{74.77}  \\ \bottomrule
    \end{tabular}
    }
\end{table}

We further evaluate the proposed methods by collaborative federated fine-tuning the learned encoder with limited labels. The results are shown in Table \ref{tab:exp_fedmae_federated_finetune}.
\textit{First}, our FedMAE outperforms all the baselines and the gap is much larger than local fine-tuning. For example, our FedMAE achieves $+13.19\%$ higher F1-score than the best-performing baseline FedSimCLR.
\textit{Second}, all the evaluated methods achieve a better performance than local fine-tuning, indicating the advantages of federated fine-tuning. This is because federated fine-tuning can use the available labeling information from all clients while local fine-tuning only uses the available labeling information of one client.

\subsubsection{Ablation Study}

\begin{table}[!htb]
	\centering
	\caption{
	Ablation study of \textbf{global and local knowledge split}. Results of federated fine-tuning are reported.
	}
	\label{tab:exp_fedmae_ablation}
	\setlength\tabcolsep{2.0pt}
	\renewcommand{\arraystretch}{1.0}
	\resizebox{1.0\linewidth}{!}{
    \begin{tabular}{lcccccccc}
    \toprule
    Sync. Parameters     & Recall & Precision & F1 & Specificity & AUC & Accuracy  \\ \midrule
    \multicolumn{7}{c}{\textit{80\% labels}} \\
    Encoder+Decoder   & 57.62 & 64.30 & 60.33 & 95.33 & 89.70 & 74.63 \\
    W/o Decoder  & 57.37 & 62.44 & 59.42 & 95.28 & 89.82 & 74.25 \\ 
    W/o Linear Projection  & 57.70 & 63.79 & 60.13 & 95.27 & 89.85 & 74.14 \\ 
    W/o Decoder Projection  & 57.71 & 63.63 & 60.02 & 95.30 & 90.06 & 74.24 \\ 
    W/o Class Token (ours)  & \textbf{58.08} & \textbf{65.76} & \textbf{61.09} & 95.32 & \textbf{90.26} & \textbf{74.77} \\ 
    \midrule
    \multicolumn{7}{c}{\textit{40\% labels}} \\
    Encoder+Decoder   & 45.97 & 51.99 & 48.40 & 94.15 & 83.35 & 68.00 \\
    W/o Decoder  & 45.61 & 51.55 & 47.89 & 94.11 & 83.57 & 67.93 \\ 
    W/o Linear Projection  & 45.99 & 51.10 & 47.98 & 94.10 & 83.28 & 67.71 \\ 
    W/o Decoder Projection  & 45.60 & 52.73 & 48.38 & 94.04 & 83.55 & 67.49 \\ 
    W/o Class Token (ours)  & \textbf{47.33} & \textbf{54.03} & \textbf{49.96} & \textbf{94.16} & \textbf{84.13} & \textbf{68.05} \\  \bottomrule
    \end{tabular}
    }
\end{table}

\textbf{Effectiveness of global and local knowledge split.}
In Section \ref{sect:knowledge_split}, we split the knowledge learned by the MAE model into global and local knowledge.
To empirically verify this hypothesis, we compare our knowledge split with different model parameters for synchronization, where the method \textit{Encoder+Decoder} synchronizes all the parameters in the MAE model, and methods \textit{W/o Decoder}, \textit{W/o Linear Projection}, \textit{W/o Decoder Projection} synchronize all parameters except for the decoder, the linear projection in the encoder's input, and the linear projection in the decoder's input, respectively. We report the diagnosis performance by using 80\% or 40\% labeled data for fine-tuning.

As shown in Table \ref{tab:exp_fedmae_ablation}, our global and local knowledge split method achieves the best performance for almost all of the metrics under 80\% or 40\% label ratios. This result verifies our hypothesis that the class embedding in the encoder of MAE contains local knowledge closely related to local data, while other parameters contain global knowledge. Synchronizing global knowledge while avoiding local knowledge achieves better performance than synchronizing all the knowledge.

\section{Conclusion}
This work aims to enable federated self-supervised learning for dermatological disease diagnosis by contrastive learning and masked autoencoders.
Based on contrastive learning, FedCLF is proposed. Devices first collaboratively pre-train a model by using decentralized unlabeled data and then fine-tune the model with limited labels.
Feature exchange is proposed to improve the data diversity for better contrastive learning on each device. 
Based on masked autoencoders, FedMAE is proposed to explore federated self-supervised learning with unlabeled data. Global and local knowledge split is developed for better knowledge aggregation and higher accuracy.
Experimental results show the effectiveness of the proposed frameworks.

\bibliographystyle{IEEEtran}
\bibliography{IEEEexample}

\end{document}